# Compact Mathematical Programs For DEC-MDPs With Structured Agent Interactions

Hala Mostafa[*]
BBN Technologies
10 Moulton Street, Cambridge, MA
hmostafa@bbn.com

Victor Lesser
Computer Science Department
University of Massachusetts, Amherst
lesser@cs.umass.edu

## Abstract

To deal with the prohibitive complexity of calculating policies in Decentralized MDPs, researchers have proposed models that exploit structured agent interactions. Settings where most agent actions are independent except for few actions that affect the transitions and/or rewards of other agents can be modeled using Event-Driven Interactions with Complex Rewards (EDI-CR). Finding the optimal joint policy can be formulated as an optimization problem. However, existing formulations are too verbose and/or lack optimality guarantees. We propose a compact Mixed Integer Linear Program formulation of EDI-CR instances. The key insight is that most action sequences of a group of agents have the same effect on a given agent. This allows us to treat these sequences similarly and use fewer variables. Experiments show that our formulation is more compact and leads to faster solution times and better solutions than existing formulations.

## 1 Introduction

Consider a robotic team dealing with a building on fire. One agent is in charge of putting out the fire, another locates and evacuates survivors and a third delivers first aid to the injured. Most of an agent's actions affect only itself (e.g. the first agent's exact approach to fire fighting and the kind of extinguisher it uses mainly affect its own progress). However, the decision-making problems of these agents are not completely independent. For example, the fire-fighting agent's decision of when to secure a given area affects how easy it will be for the rescue agent to locate survivors in that area. Decentralized Markov Decision Processes (DEC-MDPs) have been widely used to model situations like the above. However, DEC-MDPs obscure the structured agent interaction, thus missing the representational and computational savings that can be obtained due to the loose coupling among the agents.

Several models have been proposed in an attempt to exploit structured interactions, with each model catering to a different kind of structure [11, 13, 12]. Settings like the above example, where agents are largely independent except for few actions that affect the transitions and/or rewards of other agents, can be modeled using Event-Driven Interactions with Complex Rewards (EDI-CR). EDI-CR is much more compact than DEC-MDPs [9] without sacrificing expressive power.

Finding the optimal policy in decision-theoretic models can be formulated as a mathematical program. In this paper, we propose a compact formulation of EDI-CR as a Mixed Integer Linear Program (MILP). Starting with an existing non-linear formulation of general DEC-MDPs, we introduce new variables to linearize the program, but exploit structured interactions to minimize the number of variables introduced. We develop an exact formulation for the case of 2 agents and a relaxed formulation for the case of 3 or more agents.

This paper is organized as follows: the next section contains background material. Sections 3 and 4 present our 2- and 3-or-more-agents MILP formulations. Section 5 demonstrates the size and computational savings we obtain. Related work is discussed in Section 6, followed by conclusions and future work.

## 2 Background

### 2.1 EDI-CR

Event-Driven Interactions with Complex Rewards (EDI-CR) is a model developed for problems with structured transition and reward interactions [9]. It builds on the intuition that when agents are largely independent of each other, it is more natural to describe each agent's decision problem separately and list the interactions that tie these processes together. Formally, an EDI-CR instance with $n$ agents is a tuple $<\mathcal{A}, S, A, P_{1..n}, R_{1..n}, \rho, \tau, T>$ where:

- $\mathcal{A}$ is the set of $n$ agents
- $S_i$ is agent $i$'s local state space
- $A_i$ is agent $i$'s local action space
- $P_i : S_i \times A_i \times S_i \to [0, 1]$ is $i$'s local transition function
- $R_i : S_i \times A_i \times S_i \to \mathbb{R}$ is $i$'s local reward function
- $\rho = \{<(s_{k_1}, a_{k_1}), ..., (s_{k_m}, a_{k_m}), r_k>_{k=1..|\rho|}\}$ is the set of reward interactions. Each interaction involves any subset of agents and lists $r_k$, the reward/penalty of the team when the agents take the specified actions in the specified states.
- $\tau = \{<(s_{k_1}, a_{k_1}), ..., (s_{k_m}, a_{k_m}), p_k>_{k=1..|\tau|}\}$ is the set of transition interactions. The $k^{th}$ entry specifies the new transition probability $p_k$ of the state-action pair of the affected agent $k_m$ when agents $k_1$ to $k_{m-1}$ do the specified state-action pairs before $k_m$ makes its transition.
- $T$ is the time horizon of the problem

---
[*] This work was done while the first author was a Ph.D. student at University of Massachusetts, Amherst.

Individual states, actions, transitions and rewards describe each agent's decision process, while $\rho$ and $\tau$ capture the interactions among them. EDI-CR is as expressive as DEC-MDPs with local observability [9].

## 2.2 Sequence form policy representation

We use a policy representation that was first suggested by Koller et. al [7] for self-interested agents but has also been used for cooperative agents [3]. In game trees, the idea is that a policy can be characterized by the probability distribution it induces over the leaves of the tree. If two policies induce the same distribution, then they result in the same reward.

For models with local observability, a *sequence* (or *history*) of agent $i$, $s_1.a_1..s_t.a_t$, consists of $i$'s actions and local states. A history containing $T$ (time horizon) actions is a *terminal* history. For agent $i$, the set of all histories is denoted by $\mathcal{H}_i$, terminal histories by $\mathcal{Z}_i$, and non-terminal histories by $\mathcal{N}_i$. A *joint history* $h \in \mathcal{H}$ is a tuple containing one history per agent.

The *realization weight* of a history $s_1.a_1..s_t.a_t$ under a policy is the probability that the policy assigns to taking actions $a_{1..t}$ given that states $s_{1..t}$ are encountered. A history's realization weight does not include chance outcome probabilities, so it can only be 0 or 1. The vector of all realization weights will be denoted as $x$ and the weight of history $h_i$ by $x(h_i)$.

A *pure* policy deterministically chooses one action at each decision making point. In cooperative settings

Table 1: DEC-MDP as an NLP

$$\max \sum_{h \in \mathcal{Z}} \mathcal{R}(h) \prod_{g \in \mathcal{A}} x(h_g)$$
$$\text{s.t.} \sum_{a \in A_g} x(s_{g_0}.a) = 1 \qquad g \in \mathcal{A}$$
$$\sum_{a \in A_g} x(h_g.s.a) = x(h_g) \quad g \in \mathcal{A}, s \in S_g, h_g \in \mathcal{N}_g$$
$$x \in [0, 1]$$

there is at least one optimal pure joint policy, so we restrict our attention to pure policies. But even a pure policy will have multiple terminal histories with non-zero weights, because it must specify an action at each state reachable under the policy. The set of $i$'s terminal histories with weight 1 is called the policy's *support set*, $\sigma_i$, and its size is the *support size* $\|\sigma_i\|$.

In what follows, the agents are $i$, $j$ and $k$, $g$ is an arbitrary agent, and the subscript $-g$ indicates a quantity belonging to all agents but $g$.

## 2.3 Existing Mathematical Formulations

The formulation of DEC-MDP with local observability as a Non-Linear Program (NLP) is given in Table 1. In the objective function, $\mathcal{R}(h) = \beta(h)r(h)$ is the expected reward of terminal joint history $h$, where $\beta(h)$ is the probability of encountering the joint states in $h$ given the actions in $h$ and $r(h)$ is the sum of rewards of states and actions along the history. The constraints in the NLP are called *policy constraints* and guarantee that a solution to the NLP represents a legal policy where the sum of an agent's action probabilities in any state is 1. The first set of constraints ensures the sum of action probabilities at the start state is 1, while the second set ensures that the realization weights of a history's extensions add up to that history's weight.

The problem with the NLP formulation is that it results in a non-concave objective function for which no methods guarantee finding a globally optimal solution.

Aras and Dutech [3] developed a formulation for DEC-POMDPs as a MILP, thereby guaranteeing that a locally optimal solution is also globally optimal. We modify their formulation for the case of DEC-MDP with local observability. For ease of explanation, Table 2 is for the case with only 2 agents $i$ and $j$. Because the difficulty of solving a MILP increases with the number of integer variables, Aras only restricts weights of terminal histories to be integer. The constraints force the other variables to be integers as well.

To linearize the objective function, Aras introduces a *compound* variable $z_{h_i,h_j}$ for each pair of terminal his-

Table 2: DEC-MDP as a MILP

$$\max \sum_{h_i \in \mathcal{Z}_i, h_j \in \mathcal{Z}_j} \mathcal{R}(h_i, h_j) z_{h_i, h_j}$$

s.t. policy constraints and

$$\sum_{h_{-g} \in \mathcal{Z}_{-g}} z_{h_g, h_{-g}} = x(h_g) \|\sigma_{-g}\| \quad g \in \mathcal{A}, h_g \in \mathcal{Z}_g \quad (1)$$

$$\sum_{h_i \in \mathcal{Z}_i, h_j \in \mathcal{Z}_j} z_{h_i, h_j} = \|\sigma_i\| \|\sigma_j\| \quad (2)$$

$$x, z \in [0, 1] \quad x(h_g) \in \{0, 1\} \quad g \in \mathcal{A}, h_g \in \mathcal{Z}_g$$

Table 3: 2-agent EDI-CR as a MILP

$$\max \sum_{b \in B_{h_g}} \mathcal{R}_g(h_g, b) z_{h_g, b} \quad g \in \mathcal{A}, h_g \in \mathcal{Z}_g,$$

s.t. policy constraints and

$$\sum_{b \in B_{h_g}} z_{h_g, b} = x(h_g) \quad g \in \mathcal{A}, h_g \in \mathcal{Z}_g$$

$$z_{h_g, b} \leq \sum_{h_{-g} \in b} \beta(h_{-g}|h_g) x(h_{-g}) \quad g \in \mathcal{A}, h_g \in \mathcal{Z}_g, b \in B_{h_g}$$

$$x, z \in [0, 1] \quad x(h_g) \in \{0, 1\} \quad g \in \mathcal{A}, h_g \in \mathcal{Z}_g$$

tories. The variable is related to the existing $x$ variables by the identity $z_{h_i, h_j} = x(h_i) x(h_j)$.

To enforce the identity by linear constraints, Aras uses combinatorics (knowing $\|\sigma_i\|$ and $\|\sigma_j\|$) and treats the $z$ variables as counters. Constraint (1) guarantees that if $h_g$ is part of some agent $g$'s support set ($x(h) = 1$), enough compound variables involving $h_g$ are set to 1, otherwise all compound variables involving $h_g$ should be 0. Constraint (2) limits the number of compound variables that can be simultaneously set to 1.

## 3 Formulation of 2-agent EDI-CR

### 3.1 Binning histories

For the 2-agent case, the NLP in Table 1 is a Quadratic Program (QP) whose objective function has the form $\mathbf{x}^\mathsf{T} Q \mathbf{x}$ where $\mathbf{x} = [x_i, x_j]$ and $Q$ is the reward matrix. $Q(h_i, h_j) = \mathcal{R}(h_i, h_j)$ if $h_i$ and $h_j$ are terminal histories, and is 0 otherwise. The MILP in Table 2 "flattens" this matrix, multiplying each matrix entry by a compound variable created for that entry. This approach makes sense for DEC-MDPs, because agents' decision processes are tightly coupled and the rewards/transitions of one agent strongly depend on the actions taken by another. For a given history $h_i$, $\mathcal{R}(h_i, h_j)$ can vary widely depending on $h_j$, and a given row or column in $Q$ contains many distinct values, thus justifying the need for a variable per entry in $Q$.

The situation can be very different in the presence of structured interactions. An agent is only affected by a small subset of actions of another agent. So for a given $h_i$, the rewards and transition along $h_i$ do not depend on the exact actions in the history of another agent. For example, suppose $\rho$ says that agents $i$ and $j$ get a bonus of 10 if they do actions $a_1$ and $a_5$, respectively, at any point in time, and $\tau$ says that action $a_3$ of agent $j$ affects the transition probability of $a_7$ of $i$. Now suppose history $h_i \in \mathcal{Z}_i$ involves doing action $a_1$ at time 4 and $a_7$ at time 6. In this case, all histories $h_j \in \mathcal{Z}_j$ that involve doing $a_3$ before time 6 and $a_5$ any time will have the same effect on the transitions and rewards of $h_i$.

Because in EDI-CR agents have their local reward functions, we can express $Q$ as $Q_i + Q_j$. Note that this does not assume that rewards are independent; each entry in these matrices can depend on the histories of both agents. The rows (resp. columns) in $Q_i$ (resp. $Q_j$) will contain many duplicate entries, reflecting the fact that an agent is oblivious to many details of the other agent's history.

The main idea in our formulation is that for a history $h_g$, we group all histories of the other agent that have the same effect on the transitions and rewards in $h_g$ into a single **bin**. For each history $h_g$ of some agent $g$, the set of bins it induces, $B_{h_g}$, is a partition over the set of terminal histories of the other agent.

Instead of creating a variable for every pair of terminal histories, we introduce a single variable $z_{h_g, b}$ for every history $h_g$ and every bin $b \in B_{h_g}$ associated with it. In other words, we create a variable for each *distinct* entry in $Q_i$ and $Q_j$. Because structured interaction results in many duplicate entries, binning can significantly reduce the number of compound variables we introduce. Our MILP for EDI-CR is given in Table 3.

In the objective function, we fold into $\mathcal{R}_g(h_g, b)$ those quantities of $h_g$ that are oblivious to which history in $b$ is played, namely $h_g$'s transition probabilities and rewards. We therefore have $\mathcal{R}_g(h_g, b) = r_g(h_g|b)\beta(h_g|b)$. The factors on the right can be calculated using *any* history $h_{-g} \in b$

$$r_g(h_g|b) = \sum_{t=1}^{T-1} R_g(s_t, a_t, s_{t+1}) + r_\rho(h_g, h_{-g})/2$$

where $r_\rho(h, h_j)$ represents rewards (if any) that depend on actions of both agents, as specified in $\rho$. Dividing by 2 avoids double counting reward interactions. The transition probability is given by

$$\beta(h|b) = \prod_{t=1}^{T-1} P_\tau(s_{t+1}|s_t, a_t, \{a'_1..a'_t\})$$

$P_\tau$ depends on the local transition function $P_g$ and, for transitions involved in $\tau$, actions in $h_{-g}$ done up to time $t$, $\{a'_1..a'_t\}$.

We fold into $z_{h_g,b}$ quantities that depend on the particular $h_{-g}$ in the bin, namely the transition probabilities along $h_{-g}$, given history $h_g$, $\beta(h_{-g}|h_g)$. The identity defining a compound variable is therefore

$$z_{h_g,b} = x(h_g) \sum_{h_{-g} \in b} \beta(h_{-g}|h_g) x(h_{-g}) \qquad (3)$$

$z_{h_g,b}$ is the probability that $g$ plays $h_g$, multiplied by that of the other agent playing a history in $b$.

The effect of the number of interactions on the size of the formulation is clear. As the number of interactions increases, we need more bins (thus more compound variables), since each bin represents a unique way in which an agent affects another. In the extreme case of a general DEC-MDP, an agent's history needs a separate bin for each of the other's histories, essentially creating a compound variable for every pair of histories as in the DEC-MDP MILP.

### 3.2 Enforcing the identity

We need to enforce identity (3) by linear constraints. This is more challenging than in the DEC-MDP case where the binary nature of the compound variables allows the use of combinatorics to devise the constraints. In our formulation, the compound variables are not binary, and we must resort to other properties and relations to derive constraints equivalent to the identity.

Summing both sides of (3) over all bins of $h_g$ gives

$$\sum_{b \in B_{h_g}} z_{h_g,b} = x(h_g) \sum_{b \in B_{h_g}} \sum_{h_{-g} \in b} \beta(h_{-g}|h_g) x(h_{-g})$$

Since $B_{h_g}$ partitions $\mathcal{Z}_{-g}$, the double sum reduces to a sum over all histories of the other agent, giving

$$\sum_{b \in B_{h_g}} z_{h_g,b} = x(h_g) \sum_{h_{-g} \in \mathcal{Z}_{-g}} \beta(h_{-g}|h_g) x(h_{-g}) \qquad (4)$$

A legal policy prescribes an action at each state reachable by a non-terminal history with non-zero weight. As a result, histories in the support set cover all possible transitions of actions along their parents. This means that the sum of probabilities of transitions along histories in the support set must be 1. Consequently, the sum in equation (4) is 1; only the $x$s of histories in $\sigma_{-g}$ are 1, so the sum is over their corresponding $\beta$s. We therefore have the following set of constraints, one per terminal history of each agent

$$\sum_{b \in B_{h_g}} z_{h_g,b} = x(h_g) \qquad (5)$$

This constraint simply guarantees that if $h_g$ is not part of the support, all the compound variables involving $h_g$ and its bins should be 0. If $h_g$ is part of the support, it guarantees there is enough contribution from the compound variables associated with all bins of $h_g$.

However, the above constraint does not prevent one compound variable from taking too high a value at the expense of another. We can use identity (3) itself as a source of upper bounds on compound variables. Because in (3) $x(h_g)$ is either 0 or 1, we have that

$$z_{h_g,b} \leq \sum_{h_{-g} \in b} \beta(h_{-g}|h_g) x(h_{-g}) \qquad (6)$$

Together, constraints (5) and (6) strictly enforce the identity. One advantage of our constraints over the combinatorics-based constraints in the DEC-MDP formulation is that ours do not involve the size of the support set, which Aras calculates from the parameters of the problem by assuming that the number of states a state-action pair transitions to is constant. But in settings where this number depends on the particular action taken, determining the support size requires carefully looking at an agent's decision tree and the paths in it, which is non-trivial for large problems.

As for the number of constraints, the set of constraints in (5) has the same size as the constraints in the DEC-MDP MILP. The set in (6), however, is larger, because it has a constraint for each bin of each history of each agent. But as will be seen in Section 5, this does not prevent us from obtaining computational savings compared to the DEC-MDP formulation.

## 4 MILP for 3 or more agents

The idea of binning histories naturally extends beyond 2 agents. With $n$ agents, an agent's bin contains history tuples, where each tuple consists of histories of the $n-1$ other agents. Compound variables are defined by the identity

$$z_{h_g,b} = x(h_g) \sum_{h_{-g} \in b} \prod_{h_f \in h_{-g}} \beta(h_f|h, h_{-g}) x(h_f) \qquad (7)$$

As in the 2-agent case, the set of bins associated with $h_g$ is a partition over $\mathcal{Z}_{-g}$, so we can use constraint (5). The greater challenge is devising linear constraints that impose upper bounds on the $z$ variables, similar to constraint (6). With 2-agents, we simply obtained linear constraints by dropping the leading $x$ in the identity. But with 3 or more agents, doing so would result in a non-linear constraint.

In the following, we use properties of legal policies and structured interactions to derive 2 linear constraints per $z$ variable (in addition to constraint (5)) that attempt, but are not guaranteed, to enforce the identity. Note that even if the identity is violated, any feasible solution to the MILP still forms a legal set of policies, since legality is guaranteed by the policy constraints. Allowing the identity to be violated means we are solving a relaxed version of the problem.

For ease of exposition, we show the derivation of the constraints associated with a history $h_i$ of agent $i$ when $\mathcal{A}$ contains three agents $i$, $j$ and $k$.

If we assume that an action of agent $i$ can be affected by at most one other agent, we can decompose $b$ into a bin for each affecting agent such that $b = b_j \times b_k$. Dropping the leading $x$ in (7) and using the decomposition of $b$ to break down the summation gives

$$z_{h_i,b} \leq \sum_{h_j \in b_j} x(h_j) \sum_{h_k \in b_k} x(h_k)\beta(h_j|h_i,h_k)\beta(h_k|h_i,h_j) \tag{8}$$

We can obtain two linear upper bounds from the above by setting all $x(h_j)$ (resp. $x(h_k)$) to 1. But these bounds would be too loose; for a feasible solution $<x_s, z_s>$ to the MILP, $z_s$ can be very different from the $z$ calculated by applying the identity to $x_s$. In other words, the solver has too much freedom to violate the identity and set some $z$s higher than the identity allows if this improves the value of the objective function. The reward reported by the solver (the value of the objective function at $z_s$) is therefore higher than the true reward obtained when the agents follow the policies prescribed by $x_s$. The solver is optimizing a relaxed version of the problem whose optimal solution does not necessarily correspond to an optimal of the original problem. We need to tighten the upper bound so that the relaxed problem corresponds more faithfully to the original problem.

Consider the coefficient of some $x(h_j)$ in the non-linearized constraint (8):

$$\sum_{h_k \in b_k} x(h_k)\beta(h_j|h_i,h_k)\beta(h_k|h_i,h_j) \tag{9}$$

Setting all $x(h_k) = 1 \forall h_k \in b_k$ allows this coefficient to be higher than it can be under a legal policy. Regarding the coefficient as a sum over the contributions of each $h_k \in b_k$, we can decrease the coefficient by limiting the contributions of the $h_k$s. To do this, we decompose the sum in (9) into a series of sums over bins of $k$'s histories constructed from the perspective of $j$'s history $h_j$. We denote the bins of $k$'s histories induced by $h_j$ as $b_{h_j k}$ ($\bigcup b_{h_j k} = b_k$). Because $j$'s transition probability is the same under all $h_k$ in the same bin, we can factor this probability out. The coefficient can be re-written as

$$\sum_{b_{h_j k}} \beta(h_j|h_i,b_{h_j k}) \sum_{h_k \in b_{h_j k}} x(h_k)\beta(h_k|h_i,h_j)$$

Now we can use the same reasoning behind constraint (5) to get rid of $x(h_k)$ and bound the factor multiplying each $\beta(h_j|h_i,b_{h_j k})$ to be at most 1, thus restricting the coefficient of $x(h_j)$. The same can be done from the perspective of $x(h_k)$. These restrictions, together with the fact that a coefficient cannot exceed 1, allow us to approximately enforce identity (7) using constraint (5) and the following bounds

$$z_{h_i,b} \leq \sum_{h_j \in b_j} x(h_j) \left\lfloor \sum_{b_{h_j k}} \beta(h_j|h_i,b_{h_j k}) \left\lfloor \sum_{h_k \in b_{h_j k}} \beta(h_k|h_i,h_j) \right\rfloor \right\rfloor$$

$$z_{h_i,b} \leq \sum_{h_k \in b_k} x(h_k) \left\lfloor \sum_{b_{h_k j}} \beta(h_k|h_i,b_{h_k j}) \left\lfloor \sum_{h_j \in b_{h_k j}} \beta(h_j|h_i,h_k) \right\rfloor \right\rfloor \tag{10}$$

where $\lfloor x \rfloor$ denotes $\min(x,1)$.

The quest for tight linear relaxations for non-linear functions is common in the optimization literature. A trilinear term (of the form $xyz$) can be replaced by a new variable $w$ and a set of constraints that form a linear relaxation of the term's convex envelope and guarantee that $w$ is within a certain amount of the product $xyz$ [8]. Although these constraints are somewhat similar to ours, they still do not guarantee that $z_{h_i,b} > 0$ if $x(h_i) = 1$ and $j$ and $k$ play histories in $b$, which is key in bringing values of $z$ in the relaxed problem in alignment with what the identity specifies.

The idea of further binning histories within a given bin to bound the values of coefficients can be used with any number of agents. For $n$ agents, we have $n-1$ upper bounds per $z^1$ and constraint (5) can be used as-is.

## 5 Results and Discussion

### 5.1 Results of 2-agent formulations

We compare 3 formulations of EDI-CR instances: 1) the NLP formulation in Table 1, but restricted to 2 agents, 2) the DEC-MDP MILP formulation in Table 2 and 3) the EDI-CR MILP formulation in Table 3. All 3 formulations were solved using IBM ILOG Cplex [1] under the default parameters; the first using Cplex Mixed Integer QP solver, and the other two using Cplex MILP solver.

We experimented with 22 instances of the modified Mars rovers problem [9]. Each rover has a set of sites and picks the sites to visit and their order. Probabilistically, visiting a site can be slow or fast and each outcome has an associated reward. A rover visiting a site can affect the outcome probability of another rover if the latter visits a dependent site. Each such transition dependence has an entry in $\tau$ (e.g., if rover $j$ visits a site 4 before rover $i$ visits site 8, $j$ can collect samples that make it more likely that $i$ process site 8 faster). Additional reward (for complementary sites) or penalty (for redundant sites) is collected when the rovers visit certain sets of sites. Each such reward interaction has an entry in $\rho$. The number of interactions ranges from 4 to 7.

---
[1] Also, relaxations of higher order terms can be obtained by repeated application of relaxations of lower order terms [6].

We note that the time to *generate* the 3 formulations is almost the same; constructing the bins and objective function for the EDI-CR MILP is not more expensive than constructing the reward matrix for the QP or the objective function for the DEC-MDP MILP. In all 3 cases, we iterate over every pair of histories of the 2 agents to calculate their rewards and probabilities.

First, we look at behavior with respect to optimality. Even after obtaining a solution that we know is optimal (by comparing to a known optimal), Cplex may spend a long time verifying optimality. We therefore have 5 possible outcomes of a run: 1) optimal found and verified, 2) optimal found but not verified before timeout[2], 3) Locally optimal solution found, 4) Optimal not found before timeout, but a suboptimal solution was found, 5) No solution found before time out. Of our 22 instances, Table 4 compares how many fall in each category. Because a MILP solver never reports a solution that is only locally optimal, the corresponding entries are marked '-'. Our formulation resulted in a provably optimal solution in 17/22 instances. In the remaining instances, we obtained higher rewards than the other formulations, but Cplex could not verify optimality, so each of the remaining 5 instances falls into category 2 or 4. QP and DEC-MDP MILP were equally good at finding optimal solutions, although DEC-MDP MILP was better at verifying optimality. The table shows that the non-concavity of the QP can often lead the solver to just report a locally optimal solution. It also shows that in some cases, the number of compound variables introduced in the DEC-MILP is too large to allow the solver to find *any* solution before timeout (row 5).

Next, we look at the size of the MILP with and without exploiting structured interactions. We break down our 22 instances into 3 groups G1, G2 and G3 containing 5, 9 and 8 instances, respectively. We grouped instances based based on size, rather than number of interactions. The average number of interactions is 4.8 for G1, 5.3 for G2 and 5.25 for G3. Table 5 shows the number of terminal histories for each agent $|\mathcal{Z}_i|$ and $|\mathcal{Z}_j|$, the number of compound variables introduced in the DEC-MDP formulation$|z|_{DEC}$ and our EDI-CR formulation $|z|_{EDI}$, and the number of constraints (besides the policy constraints). Results were averaged over instances in each group. Clearly, the DEC-MDP formulation introduces many more compound variables than our formulation which only create as many variables as needed to distinguish between bins induced by a given history. The difference in the number of variables becomes more pronounced as the problem size increases. Although our formulation has more constraints than the DEC-MDP MILP, we next show that the increased number of constraints is offset by the large reduction in the number of variables, resulting in MILPs that are overall easier to solve.

[2]Timeout is 60 seconds for small instances and 600 seconds for larger ones.

Table 4: Optimality of 2-agent formulations

|  | QP | DEC-MDP MILP | EDI-CR MILP |
|---|---|---|---|
| 1) Optimal, Verified | 5 | 9 | 17 |
| 2) Optimal, Not verified | 9 | 5 | $x$ |
| 3) Local optimal | 5 | - | - |
| 4) Suboptimal | 3 | 6 | 5-$x$ |
| 5) No solution | 0 | 2 | 0 |

Table 5: Size of 2-agent instances and formulations

|  | $\mathcal{Z}_i$ | $\mathcal{Z}_j$ | $z_{EDI}$ | $z_{DEC}$ | $C_{EDI}$ | $C_{DEC}$ |
|---|---|---|---|---|---|---|
| G1 | 81 | 46 | 254 | 3,762 | 381 | 127 |
| G2 | 162 | 112 | 608 | 18,062 | 882 | 274 |
| G3 | 941 | 781 | 3,793 | 596,950 | 5,515 | 1,722 |

Table 6 shows the results of comparing both the time needed to find the optimal solution (reported as 'Find'), and the time needed to verify that the solution is indeed optimal (reported as 'Verify'). The times are in seconds, averaged over instances in each group. For groups where some solutions were not found/verified within reasonable time, the number of instances over which the averaging was done is indicated in brackets. In general, solving the EDI-CR MILP formulation is significantly faster than solving the other 2 formulations. There is also a large difference in the time needed to verify optimality. In G1, only 3 instances could be solved provably optimally within 60 seconds using the DEC-MDP MILP and QP formulations. In G2, the differences in time to verify optimality among the different formulations is even more pronounced. In G3, Cplex found solutions for all the instances of the EDI-CR MILP formulation, but could not verify optimality. A solution with the same quality could not be found with any of the other formulations.

Table 6: Solution time of 2-agent formulations

|  | Find EDI | Find DEC | Find QP | Verify EDI | Verify DEC | Verify QP |
|---|---|---|---|---|---|---|
| G1 | 0.29 | 8.68 | 0.57 | 0.12(3) | 3.5(3) | 0.58(3) |
| G2 | 0.59 | 10.72 | 6.4 | 0.35(6) | 21.6(6) | $> 60$ |
| G3 | 83 | N/A | N/A | N/A | N/A | N/A |

## 5.2 Results of 3-agent formulations

The 3-agent case exacerbates the problem of the DEC-MDP MILP formulation which introduces hundreds of thousands to several millions variables in our test cases. Because Cplex was unable to solve (and usually even load) the DEC-MDP MILP of our instances, we omit this formulation from further discussion.

We compare the NLP in Table 1, solved using Knitro [2] (active-set algorithm), and EDI-CR 3-agent MILP solved using Cplex [1] under the default parameters. We used 25 instances of 3-agent Mars Rovers problem broken down by size into 4 groups G1..G4 containing increasingly larger instances (sizes given in Table 7). G1 contains 10 instances and each of the other groups contains 5 instances. The number of interactions ranges from 4 to 10.

Table 7 summarizes our experimental results averaged over instances in each group. The first 4 columns show the sizes of the instances and the number of compound variables our formulation creates. The number of non-policy constraints introduced by our formulation can be calculated as $|\mathcal{Z}_i| + |\mathcal{Z}_j| + |\mathcal{Z}_g| + 2|z|$. The first 3 terms are due to constraint (5) and the last term is due to the upper bounds in (10). It is important to note that the constraint matrix, although large, is fairly sparse; the constraint of a terminal history only involves the $z$s of this history's bins, and the constraint of a $z_{h,b}$ only involves histories in $b$ of 1 affecting agent. As will be shown presently, Cplex solves EDI-CR MILPs in very little time.

To generate the NLP's objective function and construct the bins for the MILP, we need to calculate rewards for each triplet $<h_i, h_j, h_k>$. And to be able to solve large instances like the ones we experiment with, we need to identify and remove dominated histories as in [3][3]. The times for these 2 steps are in the first 2 columns of the 'Times' section of Table 7. Although expensive, the first step is unavoidable and the second saves time and space for both NLP and MILP. Column 'Bin' shows the time to generate constraints (5) and (10) for the MILP. Although this step is rather expensive, it results in a MILP that is solved at least an order of magnitude faster than the NLP (shown in the 'Sol.' columns). Even with our current implementation (unoptimized for speed), the time to solve the NLP far exceeds the constraints generation time and the MILP solving time combined. The difference becomes more pronounced with larger instances (for G4, we timed out Knitro and report the reward obtained after 30 minutes).

---
[3]None of the formulations would otherwise fit in memory. Reported numbers are those of undominated histories.

Finally, Table 7 compares the rewards obtained by the policies from the NLP and MILP solutions. As explained in Section 4, our MILP is a relaxation of the original problem where some $z$s can be higher than their values under identity (7). The solution reported by the MILP is therefore an over-estimate of the optimal reward. The table shows the reward of the MILP and NLP policies as a percentage of this over-estimate. For smaller instances, MILP and NLP give comparable rewards, but on larger ones, Knitro is unable to produce good policies within 30 minutes.

Whereas we do not know of a way to improve the quality of the NLP solution, the MILP can be improved by improving the correspondence between the MILP and the original problem, i.e. making the upper bounds on $z$s tighter and/or obtaining lower bounds on $z$. This would result in a space of feasible solutions that better resembles that of the original problem, and would prevent the solver from pursuing solutions that appear to be good, but are sub-optimal in the original space.

## 6 Related Work

Researchers have formulated decision problems as mathematical programs with the aim of benefiting from available industrial-grade optimization packages. Besides formulations for the general DEC-MDPs [3], special cases were also addressed. Transition-Independent DEC-MDPs [5] were formulated as MILP [14], but this formulation cannot be useful for EDI-CR because transition dependence adds significant complexity (TI-DEC-MDP is NP-complete while EDI-CR is NEXP-complete) and changes the way we approach the problem. Aras et. al [4] developed a QP for Network Distributed POMDPs (ND-POMDP) [10], a model where agents either do not interact, or interact very closely (coarse-grained independence). In EDI-CR, interactions are specified at the level of actions, and we exploit this fine-grained specification.

Several variants and sub-classes of DEC-(PO)MDPs have been proposed, each catering to a different kind of structured interaction. For example, IDMG and DPCL [11, 12] address settings where interaction among agents is specific to a set of interaction states. We cannot (easily) represent situations where some agents' actions affect outcomes of other actions using these models because their agents are assumed to be tightly coupled in the interaction states, so all of an agent's actions taken in those states affect the other agent. Moreover, IDMG's solution approach assumes knowledge of the joint state (through free communication) when the agents are in the interaction states, an assumption not in EDI-CR. TD-POMDP [13] is a model somewhat similar to EDI-CR, but its solution approach is very different and has not been demonstrated in situations where all agents affect each other.

Table 7: Comparison of 3-agent formulations

| | Size | | | | Time (s) | | | | | | | % Reward | |
|---|---|---|---|---|---|---|---|---|---|---|---|---|---|
| | $\mathcal{Z}_i$ | $\mathcal{Z}_j$ | $\mathcal{Z}_k$ | $z_{EDI}$ | Reward Calc. | Dom. Removal | Bin | MILP Sol. | NLP Sol. | MILP Total | NLP Total | MILP | NLP |
| G1 | 116 | 114 | 102 | 507 | 192 | 78 | 6.6 | 1.2 | 27.4 | 278 | 297 | 88% | 86% |
| G2 | 163 | 170 | 220 | 898 | 1296 | 353 | 56.2 | 7.22 | 284.9 | 1712 | 1934 | 85.7% | 88.3% |
| G3 | 285 | 184 | 158 | 1265 | 2063 | 802 | 119 | 38.3 | 588 | 3022 | 3916 | 88.7% | 87.2 % |
| G4 | 272 | 229 | 608 | 1555 | 3032 | 483 | 413 | 68.4 | 1800 | 3996 | 5315 | 86.2% | 51.5% |

## 7 Conclusion

This paper presents compact MILP formulations of a class of DEC-MDPs where there are structured transition and reward interactions among agents. Previously, this problem would be formulated as an NLP, which is expensive to solve and can result in suboptimal solutions, or using a general MILP formulation for DEC-MDPs, whose size is typically prohibitive. Our formulation successfully exploits structured interactions using the insight that most action histories of a group of agents have the same effect on a given agent, thereby allowing us to treat these histories similarly and use fewer variables in the formulation. Experiments show that our MILP is more compact and leads to faster solution times and generally better solutions than formulations ignoring the structure of interactions. Our formulation allows us to solve larger problems which would otherwise be intractable.

One future direction for this work involves approximately binning where histories whose effects are similar enough are grouped in the same bin, thus reducing the number of compound variables needed.